\pdfoutput=1

\documentclass[11pt]{article}

\usepackage[final]{acl}

\usepackage{times}
\usepackage{latexsym}

\usepackage[T1]{fontenc}

\usepackage[utf8]{inputenc}

\usepackage{microtype}

\usepackage{inconsolata}

\usepackage{graphicx}
\usepackage{subcaption}
\usepackage{gb4e} 
\usepackage{enumerate}
\usepackage{multirow}
\usepackage{float}
\usepackage{amsmath} 

\usepackage{graphicx}
\usepackage{dblfloatfix} 
\usepackage{caption}
\usepackage[subrefformat=parens]{subcaption}
\usepackage{booktabs}
\usepackage{longtable}
\usepackage{tabularray}

\title{A Dual-Task Paradigm to Investigate Sentence Comprehension Strategies in Language Models}

\author{Rei Emura \\
  Tohoku University \\
  \texttt{rei.emura.r4@dc.tohoku.ac.jp} \\\And
  Saku Sugawara \\
  National Institute of Informatics \\
  The University of Tokyo \\
  \texttt{saku@nii.ac.jp} \\}

\begin{document}
\maketitle 
\begin{abstract}
Language models (LMs) behave more like humans when their cognitive resources are restricted, particularly in predicting sentence processing costs such as reading times. However, it remains unclear whether such constraints similarly affect sentence comprehension strategies. Besides, existing methods do not directly target the balance between memory storage and sentence processing, which is central to human working memory. To address this issue, we propose a dual-task paradigm that combines an arithmetic computation task with a sentence comprehension task, such as ``The 2 cocktail + blended 3 =...'' Our experiments show that under dual-task conditions, GPT-4o, o3-mini, and o4-mini shift toward plausibility-based comprehension, mirroring humans' rational inference. Specifically, these models show a greater accuracy gap between plausible sentences (e.g., ``The cocktail was blended by the bartender'') and implausible sentences (e.g., ``The bartender was blended by the cocktail'') in the dual-task condition compared to the single-task conditions. These findings suggest that constraints on the balance between memory and processing resources promote rational inference in LMs. More broadly, they support the view that human-like sentence comprehension fundamentally arises from the allocation of limited cognitive resources.
\end{abstract}

\section{Introduction}
Working memory is a cognitive system that temporarily stores and maintains information necessary for processing in an accessible state \cite{Atkinson1971Control, Baddeley1974Working, Baddeley2003Working}. It is essential for understanding language in humans \cite{Just1992Capacity}. 

Comparing the working memory of LMs with that of humans helps us understand what makes sentence comprehension more human-like.
The limitation of cognitive resources (analogous to working memory in humans) for sentence comprehension makes LMs behave more like humans \cite{Futrell2020Lossy,Hahn2022Resource,Kuribayashi2021Lower,Kuribayashi2022Context,Oh2022Comparison,Oh2023Why,Timkey2023Language,Wilcox2025Bigger}. Specifically, LMs with restricted memory resources better approximate human reading costs, such as reading times. These findings suggest that resource constraints may be a fundamental property of human-like language comprehension.

\begin{figure}[t]
    \includegraphics[width=\columnwidth]{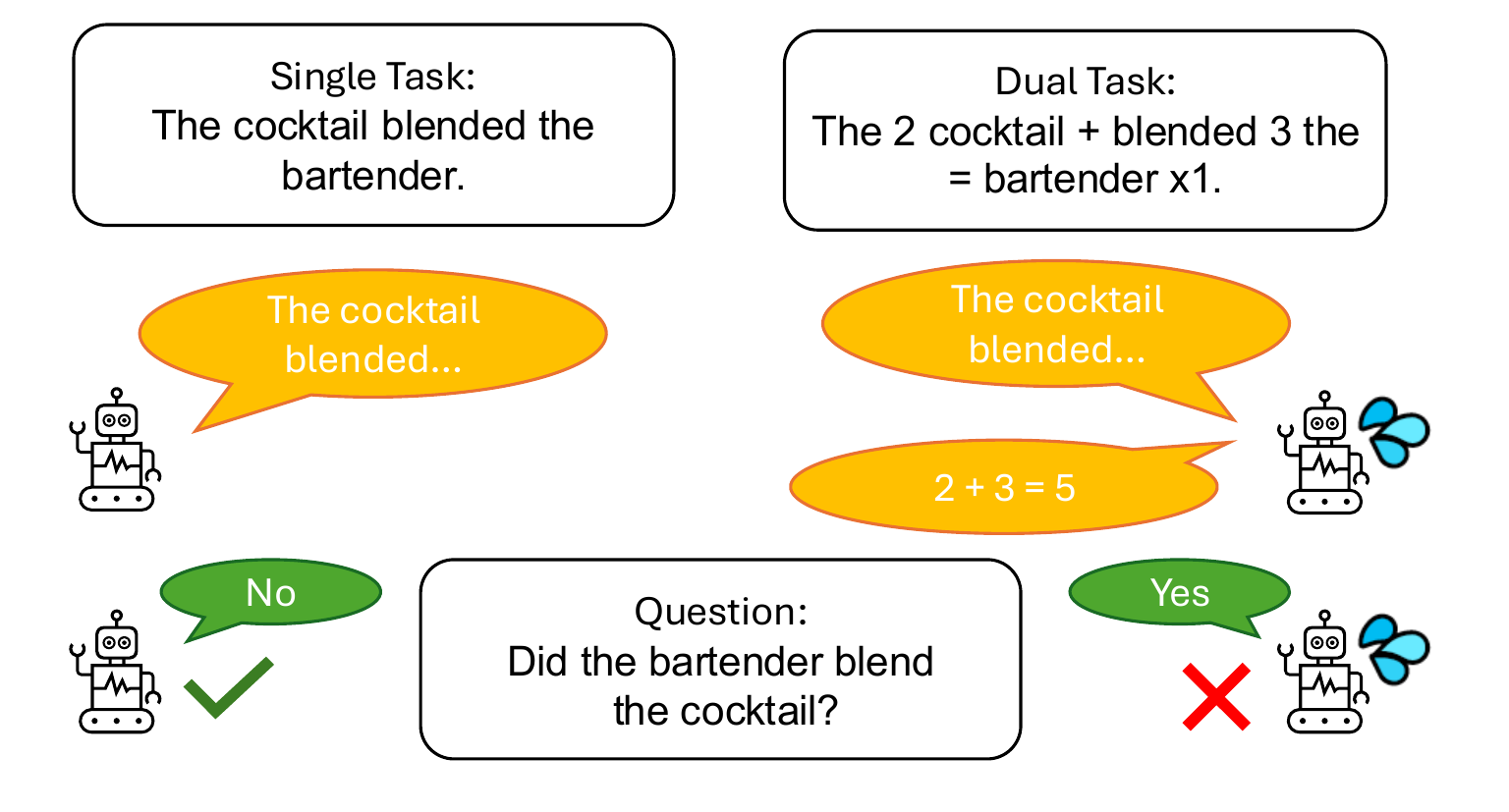}
    \caption{Overview of hypothesis and tasks. This study investigates whether language models, similar to humans, prioritize plausibility over grammar when understanding implausible sentences in a dual-task situation where they simultaneously perform calculations and sentence comprehension.}
    \label{fig:model}
\end{figure}

However, it remains unclear whether LMs' reading strategies exhibit patterns analogous to those of humans under limited cognitive resources. 
Previous studies have shown that LMs achieve lower accuracy on complex sentences, suggesting that LMs rely on working-memory-like mechanisms during sentence processing \cite{Amonyal2025Comparing,Amouyal2025When,Irwin2023Bert}. 
We therefore examine whether LMs adopt human-like comprehension strategies when cognitive resources are constrained.

Existing approaches to constraining cognitive resources in LMs also face methodological limitations. Prior work has mainly manipulated input length or model parameters \cite{Asami2024What,Kuribayashi2021Lower,Kuribayashi2022Context,Oh2022Comparison,Oh2023Why,Timkey2023Language,Wilcox2025Bigger}. These methods either fail to capture the balance between storage and processing that characterizes human working memory or fail to induce modifications within a single LM.

To address the methodological issue, we propose a dual-task paradigm in which models simultaneously solve arithmetic problems and answer comprehension questions, such as ``The 2 cocktail + blended 3 =...'' (see Figure \ref{fig:model}).
We compare comprehension accuracy across three conditions: (i) single task (comprehension of sentences without calculation), (ii) noisy single task (comprehension of sentences with embedded calculation, but no concurrent arithmetic solving), and (iii) dual task (comprehension of sentences with embedded calculation while solving the arithmetic problems). 

We focus on one characteristic property of human reading strategies: limited cognitive resources promote rational inference \cite{Futrell2017L2,Gibson2013Rational,Gibson2016Rational}.
Rational inference involves interpreting semantically implausible sentences (e.g., ``The cocktail blended the bartender.'') as plausible meanings consistent with world knowledge (e.g., ``The bartender blended the cocktail.''), prioritizing prior knowledge about plausibility over grammatical structure.
Humans are more likely to adopt this strategy under high cognitive load, than under low load \cite{Ayasse2021Principle,Ferreira2003Misinterpretation,Gibson2013Rational}.\footnote{The tendency to prioritize plausibility over grammatical information has been explained not only by rational inference theory, but also by the good-enough theory \cite{Christianson2016,Ferreira2007Good} and shallow processing \cite{Sanford2002Depth}. However, we focus on the observed behavioral patterns, and therefore do not engage with their underlying mechanisms.}

Taken together, we examine whether LMs adopt rational strategies under dual-task conditions and, if so, which conditions promote such strategies.\footnote{Our data, codes, and results are available at \url{https://github.com/reiemura/llm-dual-task}.} Our contributions are summarized as follows:
\begin{enumerate}[(i)]
    \item We propose a dual-task paradigm to test the behavior of LMs under limited cognitive resources. This paradigm allows us to observe the LMs' function in integrating memory with sentence processing, analogous to human working memory. 
    \item We demonstrate that GPT-4o, o3-mini, and o4-mini shift their comprehension strategies toward rational inference under the dual-task condition, showing a larger accuracy gap between plausible and implausible sentences than in the single-task and noisy single-task conditions.
    \item We show that these models are more likely to misunderstand implausible sentences with passive, dative, or benefactive structures. This suggests that they rely more on world knowledge and superficial word order than on function words under limited cognitive resources.
    \item Our findings support the view that resource constraints are a fundamental property of human-like language comprehension, extending previous evidence from reading costs to reading strategies.
\end{enumerate}



\section{Related Work}
\subsection{Approach to Constrain the LMs' Working Memory}
There are two major approaches in computational psycholinguistics for constraining the cognitive resources of LMs: manipulating the input text and altering the models' parameters.

For the first approach, \citet{Asami2024What} manipulates the length of entire sentences and compares the accuracy differences between plausible and implausible sentences. Their results show that longer sentences reduce accuracy for both types, indicating that increased length does not promote greater reliance on plausibility information. This may be because such manipulation merely increases processing demands without engaging the balance between storage and processing, which is a central feature of human working memory. A key function of working memory is its dual role of storage and processing, such as remembering a series of numbers while performing a distracting task \cite{Atkinson1971Control, Baddeley1974Working, Baddeley2003Working,Just1992Capacity}.

Regarding the second approach, studies have shown that reducing the number of attention heads \cite{Timkey2023Language}, larger perplexity \cite{Kuribayashi2021Lower,Oh2022Comparison,Oh2023Why}, limiting context access \cite{Kuribayashi2022Context}, and small training data size and training steps \cite{Wilcox2025Bigger} lead to better prediction of human reading times. However, these manipulations create different model configurations, each trained for specific tasks, and therefore reflect differences between models rather than changes within a single model. This is analogous to comparing human participants with different working memory capacities, rather than examining how one participant adapts under varying conditions.

To address these limitations, we introduce a dual-task paradigm in which arithmetic expressions are interleaved with sentence words (see Figure \ref{fig:model}). This design maintains the need for memory storage while imposing additional processing demands, thereby constraining the working-memory function that balances storage and processing. Furthermore, by manipulating the task rather than the model, our approach sheds light on how a single model alters its reading strategy under resource constraints. 

\subsection{Dual-Task Approaches in LMs and Humans}
Previous work on multi-task processing in LMs has primarily aimed to improve performance or efficiency by enabling models to handle multiple tasks simultaneously \cite{Cheng2023Batch, Son2024Multi}. These studies focus on optimizing accuracy or speed and are not designed to investigate how cognitive resource limitations affect language comprehension.

In contrast, some studies have attempted to constrain LMs’ working memory using n-back tasks \cite{Kirchner1958Age}, with the explicit goal of taxing internal memory resources \cite{Gong2024Working,Zhang2024Working}. While these studies share our objective of probing working memory limitations, n-back tasks primarily engage numerical memory and calculation rather than sentence comprehension.

We build on this latter approach by focusing on working memory in language comprehension. 
Specifically, our dual-task approach is inspired by human working-memory paradigms, particularly the operation span task \cite{Turner1989Working}. 
This task requires participants to perform arithmetic operations while memorizing words or letters. In the original version \cite{Turner1989Working}, a mathematical problem followed by a to-be-remembered word is presented, such as ``(3 × 4) + 11 = 20?    BEAR.'' Participants first read the problem aloud and judge whether the answer is correct, then read and memorize the following word. After several trials (typically two to six), they are asked to recall the memorized words in the correct order. 

\section{Methods}
\subsection{Task}
We conduct three types of question-answering tasks. The Dual Task is designed according to the LM’s specifications (where a list of strings is more suitable). The Noisy Single Task is included to examine whether performance changes are due to the presence of noisy arithmetic expressions or to the additional cognitive demands of the Dual Task. Exact prompts are in Appendix \ref{sec:prompts}.

\begin{enumerate}[(i)]
\item \textbf{Single Task (Single)}: The LM receives a sentence without any embedded arithmetic problems, then answers a comprehension question about the sentence. 
\item \textbf{Noisy Single Task (Noisy)}: The LM receives a sentence with embedded arithmetic problems but ignores them, then answers a comprehension question about the sentence. 
\item \textbf{Dual Task (Dual)}: The LM receives a sentence with embedded arithmetic problems, solves the arithmetic problems, and then answers a comprehension question about the sentence.
\end{enumerate}

\subsection{Dataset}
\begin{table*}[htb]
\centering
\begin{tabular}{p{2.5cm}p{2.5cm}p{9cm}}
\toprule
Factor & \multicolumn{1}{l}{Variable} & Premise and stimuli (Implausible except for the top) \\ \midrule
\multirow{2}{*}{Plausibility} & Plausible  & The bartender blended the cocktail. (Premise) \\
                              & Implausible  & The cocktail blended the bartender. (Premise) \\ \midrule
\multirow{6}{*}{Construction}                  & Transitive  & The cocktail blended the bartender. (Premise)   \\
                              & Passive & The bartender was blended by the cocktail. (Premise) \\
                              & Dative   & The chef sent the friend to the gift. (Premise)\\
                              & Exp.Subj. & The view missed the traveler. (Premise)\\
                              & Exp.Obj.   & The researcher encouraged the results. (Premise)\\
                              & Ben.For & The uncle bought the nephew for the toy. (Premise)\\ \midrule
\multirow{5}{*}{Task} &
  Single &
  The cocktail blended the bartender and the intruder cited the patent after the neurologist baffled the hippie. (Stimuli)\\
 &
  Noisy \& Dual &
  The 5 cocktail + blended 6 the = bartender x5633 and 9 the + authorities 3 agitated = the x5634 organist 6 after + the 8 infantryman = saluted x5635 the 3 pollster. (Stimuli, 1dig.2add.)
 \\ \midrule
Correct Answer & Yes / No  & Did the bartender blend the cocktail? (Question) \\ \bottomrule
\end{tabular}
\caption{Examples of premises and stimuli depending on factors and variables. Plausibility and Construction display premises, and Task displays stimuli we actually used in the experiment.
Abbreviations: Exp.Subj = Experiencer Subject; Exp.Obj. = Experiencer Object; Ben.For = Benefative For.}
\label{tab:stimuli}
\end{table*}

We use a subset of stimuli from the GELP dataset \cite{Asami2024What}. The dataset consists of sentence–question pairs. Each sentence includes one premise connected with two propositions.\footnote{GELP also contains sentences with one or no propositions. We use only sentences with two propositions, corresponding to the high memory-load condition.}

Premises are manipulated by plausibility (\textbf{Plausible} / \textbf{Implausible}) and construction (\textbf{Transitive} / \textbf{Passive} / \textbf{Dative} / Experiencer Subject (\textbf{Exp.Subj.}) / Experiencer Object ({\textbf{Exp.Obj.})  / Benefactive For (\textbf{Ben.For})), as illustrated in Table~\ref{tab:stimuli}. 
Although the original dataset includes eight constructions, two are excluded during preprocessing (see Section \ref{sec:preprocessing} for details). 

In addition, to conduct the Dual Task and Noisy Single Task, we add arithmetic expressions to these sentences. Randomly generated computation problems are inserted with identifiers (``x1,'' ``x2,'' ``x3'' ...). After the ``=,'' the corresponding identifier string (``x1,'' ``x2,'' ``x3''...) is appended. 
Each arithmetic expression is interleaved into the sentence one word at a time. If the end of the sentence is reached in the middle of an arithmetic problem, the remaining calculation problems are not added. 

We use ten types of arithmetic problems, varying in both digit length (1, 3, 5, 10, and 30 digits) and the number of addends (two vs. three). They are abbreviated as Xdig.Yadd. (e.g., 1dig.2add. indicates the addition of two one-digit numbers). Example stimuli for some arithmetic types are provided in Appendix~\ref{sec:stimulibyarith}.

All comprehension questions are binary (Yes/No), balanced such that half of the correct answers are Yes and half are No.
The final dataset contains 2,560 sentence–question pairs ($2$ plausibility levels $\times$ $8$ constructions $\times$ $160$ items).

\section{Experiments}
\subsection{Experimental Setup}
We evaluate seven LMs:  GPT-4o \cite{Openai2024GPT-4}, o3-mini, o4-mini,\footnote{https://openai.com/index/introducing-o3-and-o4-mini/} GPT-4.1 \cite{Openai2024GPT-4}, DeepSeek-V3 \cite{DeepSeek2025DeepSeek-V3}, Llama-3.3 \cite{Grattafiori2024Llama3}, and Gemma-3 \cite{Gemma2025Gemma3}. 
The models and prompts are selected based on the following two criteria: (\mathrm{i}) accuracy for the implausible condition in the Single Task must be at least 70\%, and (\mathrm{ii}) accuracy for the arithmetic problems in the Dual Task of the 1dig.2add. condition must be at least 80\%.\footnote{We also test GPT-3.5-turbo (\url{https://platform.openai.com/docs/models}), but it does not meet these criteria.} We set the temperature to 0.0.\footnote{This operation was restricted for the o3-mini and o4-mini.}

\subsection{Evaluation Metrics} 
\subsubsection{Preprocessing}\label{sec:preprocessing}
Prior to analysis, we filter the data based on model performance on the Single-Task comprehension task and the Dual-Task arithmetic problems.

First, we compute single-task accuracy for each construction and plausibility condition and exclude those below 80\%. As a result, two constructions (double object and benefactive double object) are excluded for all models, with one additional construction excluded for DeepSeek-V3 and three for Gemma-3. This ensures that analyses include only constructions that models reliably comprehend in the single task.

Second, we exclude arithmetic problem types with incorrect answers exceeding 40\%. 
We also remove trials where the arithmetic problem was solved incorrectly or the comprehension response could not be extracted.  
This filtering step ensures that the models analyzed do not adopt strategies that ignore or skip the arithmetic task.

\subsubsection{Accuracy of Comprehension Task}
We statistically analyze whether the plausibility effect is larger in the dual task than in the single task and noisy single task, using \texttt{R}  \cite{RCoreTeam2025R}. We use a per-item, non-parametric difference-in-differences procedure. 
For each item $i$ and task $t$, we compute the mean accuracy  $\hat{p}_{itp}$ within each plausibility level $p$. The within-task plausibility contrast is defined as:
\begin{equation}
    \Delta_{it} = \hat{p}_{it,\text{Plausible}} - \hat{p}_{it,\text{Implausible}}.
\end{equation}

For each item, we then calculate two difference-in-differences contrasts:
\begin{equation}
    D^{\sc{DS}}_i = \Delta_{it,\text{Dual}} - \Delta_{it,\text{Single}} 
\end{equation}
   \begin{equation} 
    D^{\sc{DN}}_i = \Delta_{it,\text{Dual}} - \Delta_{it,\text{Noisy}}.
\end{equation}

Finally, we conduct one-sided Wilcoxon signed-rank tests to assess $H_0: median(D) \leq 0$ separately for $D^{\sc{DS}}_i$ and $D^{\sc{DN}}_i$. The null hypothesis $H_0$ is rejected if the one-sided test is significant at $\alpha = 0.05$.


\subsection{Human Experiment}
We evaluate LMs against well-established human phenomena following prior work in psycholinguistics \cite{Ayasse2021Principle,Ferreira2003Misinterpretation,Gibson2013Rational}.
Nevertheless, to verify whether consistent patterns can be replicated in our dual-task paradigm, we collected a small dataset of human data.

We recruit 33 native English speakers using a crowdsourcing service called Prolific (\url{https://www.prolific.com/}). We only recruited people who live in the U.S., the U.K., Ireland, Australia, or New Zealand, have a bachelor's degree (this is because this dual-task is somewhat difficult to follow), and have approval rates of 97\% or higher.  
We obtained informed consent prior to the experiment, and compensated the participants approximately £10.00 for their 1-hour participation. 

The design, stimuli, and procedures are the same as the LM experiment, but the arithmetic is limited to 1dig.2add. because other arithmetic types are too challenging for humans. 
The stimuli include 10\% of the items for LMs and are distributed into 12 lists using the Latin square design. Each list contains 65 trials, and each item appears only once in one condition in each list. 
The appendix~\ref{HumExpProc} provides details of the procedures and instructions for participants. 

In data analysis, we adopted the same criteria as in the LM analysis for participant-level screening, sentence-construction screening, and trial-level screening. After applying these criteria, data from 14 participants (mean age $\pm$ standard deviation: 39.07 $\pm$ 12.30; 9 females and 5 males) are retained for analysis. In the sentence-construction screening, the double object construction and benefactive double object construction are excluded, consistent with the LM analysis. Ben.For and Exp.Obj. are also excluded from the human analysis. We do not apply statistical tests for human data because the sample size is too small.

\subsection{Results}\label{sec:results}
\begin{figure*}[htb]
  \centering
  \includegraphics[width=\textwidth,keepaspectratio]{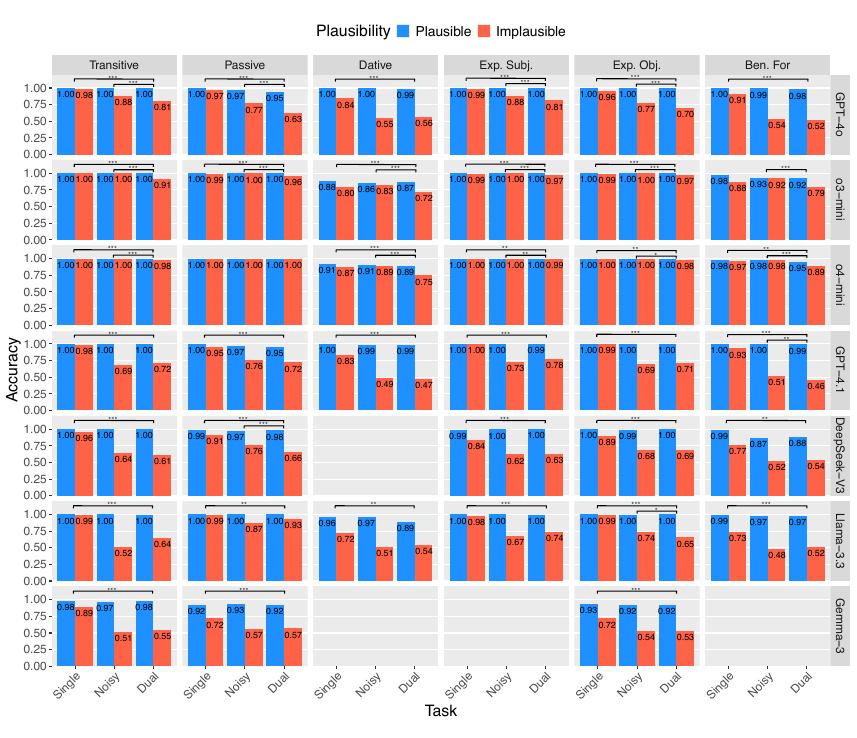}
  \caption{Mean accuracy of comprehension tasks by plausibility, task, LM, and construction. GPT-4o is likely to show significantly larger accuracy drops for implausible sentences in the dual task than in the single and noisy single tasks. o3-mini and o4-mini are likely to show a similar pattern but maintain near-ceiling accuracy across conditions. Other models show reduced accuracy in both noisy single and dual tasks, with no clear difference between them. Some conditions are excluded during preprocessing (see Section \ref{sec:preprocessing}).
  *\textit{p} < 0.05. **\textit{p} < 0.01. ***\textit{p} < 0.001. Abbreviations: Single = Single Task; Noisy = Noisy Single Task; Dual = Dual Task; Exp.Subj = Experiencer Subject; Exp.Obj. = Experiencer Object; Ben.For = Benefative For.}
  \label{fig:graphbycond}
\end{figure*}

Figure~\ref{fig:graphbycond} shows the mean comprehension accuracy by plausibility, LM, and construction.  
As seen in the graph, whether the Dual Task promotes rational inference depends on both the LM and the sentence construction. 
The models can be grouped into three categories as follows.   
\begin{enumerate}[(i)]
\item GPT-4o is likely to use rational inference in the dual task. Four out of six constructions show lower accuracy for implausible sentences in the dual task than in either the single or noisy single tasks.
\item o3-mini and o4-mini show a similar tendency but maintain high accuracy across all conditions (around 100\%).  Four out of six constructions show significantly lower accuracy for implausible sentences in the dual task than in the single or noisy single tasks.  
\item The other models, i.e., GPT-4.1, DeepSeek-V3, Llama-3.3, and Gemma-3, are likely to rely on rational inference in both the noisy single and dual tasks. These models generally show significantly lower accuracy in the noisy single and dual tasks than in the single task, but no significant difference between the noisy single and dual tasks.
\end{enumerate}


In summary, the results suggest that GPT-4o, o3-mini, and o4-mini are more likely to engage in rational inference when cognitive resources are constrained. A consistent trend is observed when analyzing plausibility effects across different arithmetic problems (see Appendix~\ref{sec:accbyarith}) and correct answers (see Section~\ref{sec:accbyanswer}).


Table \ref{tab:humandata} shows results from the human experiment.
The dual task yields generally lower accuracy and a larger difference between plausible and implausible sentences than the single and noisy single tasks.  
Specifically, the mean accuracy difference between plausible and implausible conditions is 0.064 in the dual task, compared with 0.025 in the single task and 0.035 in the noisy single task.
This pattern suggests increased reliance on plausibility under dual-task conditions in humans. 
It parallels prior psycholinguistic findings \cite{Ayasse2021Principle,Ferreira2003Misinterpretation,Gibson2013Rational}, and patterns observed in our data for several LMs, including GPT-4o, o3-mini, and o4-mini.

\begin{table}[]
\centering
\small
\begin{tabular}{llll}
\toprule
            & Single & Noisy & Dual   \\ \midrule
Plausible   & 0.95 (0.23) & 0.88 (0.33)       & 0.82 (0.39) \\
Implausible & 0.92 (0.27) & 0.84 (0.37)       & 0.75 (0.43) \\ \bottomrule
\end{tabular}
\caption{Mean accuracy (with standard deviations) in the human experiment. The dual task shows a generally lower accuracy and a larger plausible–implausible accuracy difference than the single and noisy single tasks. Abbreviations: Single = Single Task; Noisy = Noisy Single Task; Dual = Dual Task.}
\label{tab:humandata}
\end{table}

\section{Analysis}
\subsection{Effects of Plausibility by Correct Answer}\label{sec:accbyanswer}

Figure \ref{fig:graphbyanswer} presents the mean accuracy rates of comprehension questions by plausibility, task, LM, and correct answer (Yes or No). When the correct answer is ``Yes,'' all models show larger plausibility contrasts under the dual task condition than in the single or noisy single tasks. This effect is driven by a substantial decrease in accuracy for implausible sentences under the dual task. That is, the models often fail to correctly respond ``Yes'' when asked whether an implausible sentence expressed an implausible meaning, instead responding ``No.'' 

On the other hand, when the correct answer is ``No,'' GPT-4o, o3-mini, and o4-mini still exhibit tendencies consistent with rational inference, similar to the Yes condition. Other models do not show such a pattern. 
In summary, consistent with the results in Section~\ref{sec:results}, GPT-4o, o3-mini, and o4-mini reliably demonstrate a shift toward rational inference across both answer types.

\begin{figure}[h!]
  \begin{center}
  \includegraphics[width=\columnwidth]{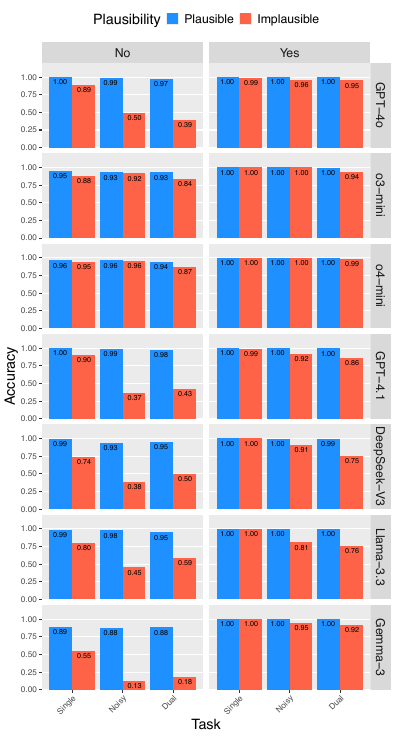}
  \caption{Mean accuracy of comprehension tasks by plausibility, task, LM, and correct answer. When the correct answer is “Yes,” all models show larger plausibility contrasts in the dual task, driven by reduced accuracy for implausible sentences. When the correct answer is “No,” only GPT-4o, o3-mini, and o4-mini show a similar pattern. Abbreviations: Single = Single Task; Noisy = Noisy Single Task; Dual = Dual Task.}
  \label{fig:graphbyanswer}
  \end{center}
\end{figure}

\subsection{Conditions Where the Implausible Sentences are Misunderstood}

\begin{figure*}[h]
  \begin{center}
  \includegraphics[width=\textwidth]{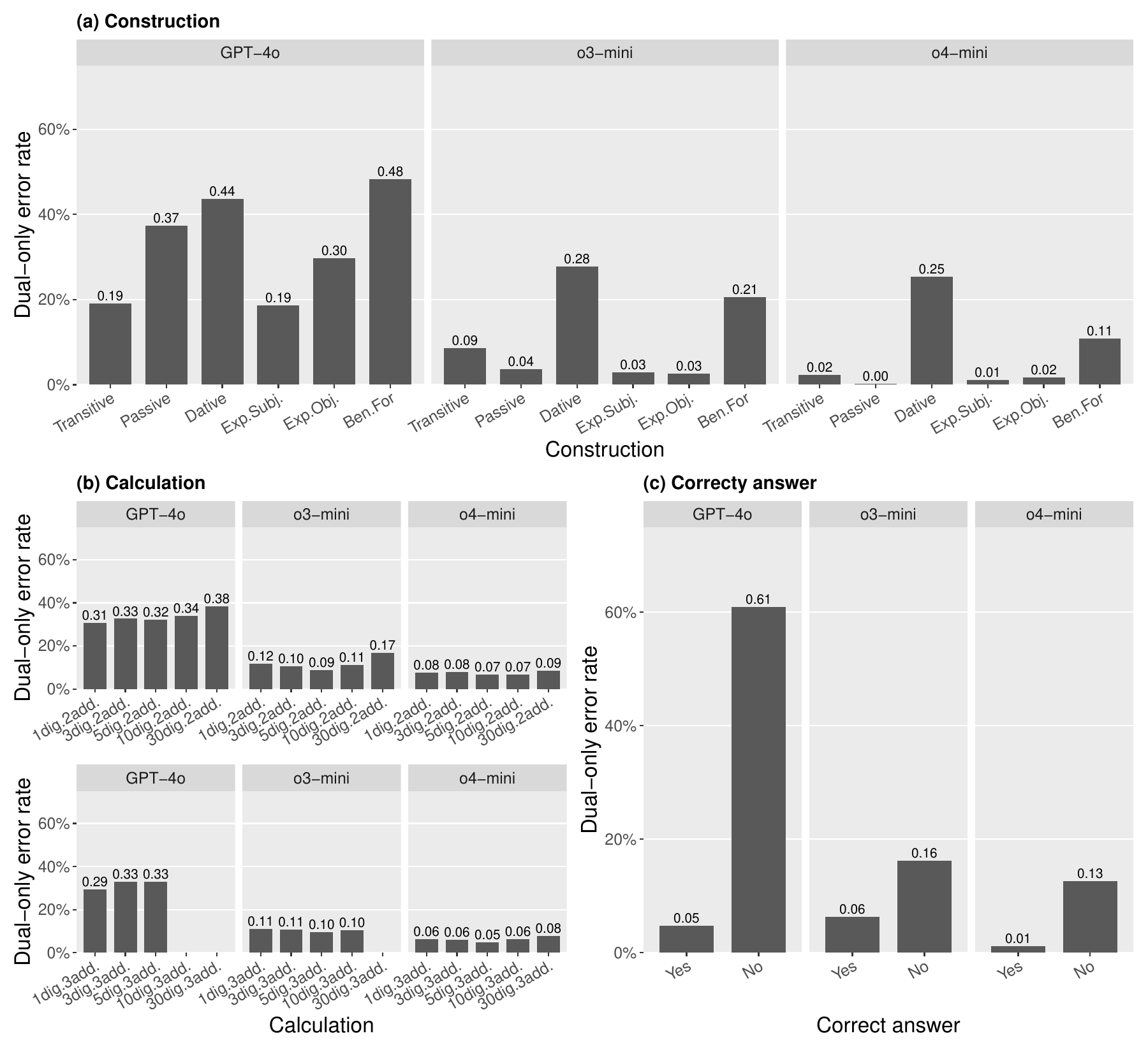}
  \caption{Proportion of implausible items answered correctly in the single and noisy single tasks but incorrectly in the dual task, by construction, arithmetic condition, and correct answer. (a) Errors are most frequent for dative and benefactive-for constructions (and passive for GPT-4o). (b) Errors peak in the 30-digit condition. (c) Errors are more frequent for “No” answers. Some conditions are excluded during preprocessing (see Section \ref{sec:preprocessing}).Abbreviations: Exp.Subj = Experiencer Subject; Exp.Obj. = Experiencer Object; Ben.For = Benefative For.}
  \label{fig:graphimp}
  \end{center}
\end{figure*}

The previous sections show that GPT-4o, o3-mini, and o4-mini tend to shift toward rational inference under the dual task. Here, we examine when these models are most likely to misinterpret implausible sentences as plausible under the dual task. Figure~\ref{fig:graphimp} illustrates the proportion of implausible items that are answered correctly in the single and noisy single tasks but incorrectly in the dual task. 

Examining the distribution across constructions in Figure~\ref{fig:graphimp} (a), GPT-4o, o3-mini, and o4-mini show the highest error rates for dative and benefactive-for constructions. GPT-4o also frequently fails on passive sentences. These constructions share a key property: when function words (and morphemes) are removed, the remaining word sequence appears semantically plausible. For instance, removing function words from ``The bartender was blended by the cocktail'' and ``The chef sent the friend to the gift'' yields ``bartender blend cocktail'' and ``chef send friend gift'' respectively, which could be interpreted as plausible events. Therefore, this suggests that under resource constraints, these models rely primarily on the superficial word sequence of content words and plausibility derived from world knowledge, rather than function words.

Next, the distribution across the correct answer in Figure~\ref{fig:graphimp} (c) shows that errors are more frequent when the correct answer is ``No'' than when it is ``Yes.''  
This indicates that when asked comprehension questions such as ``Did the bartender blend the cocktail?'' the models tend to respond ``Yes,'' showing a bias toward affirmative answers. 
This pattern resembles acquiescence, called  ``yea-saying,'' observed in humans during Yes/No question answering tasks \cite{Jackson1958Content,Knowles1999Why}. LM's acquiescence bias has also been observed \cite{Dentella2023Systematic}.

Finally, the calculation graph in Figure~\ref{fig:graphimp} (b) shows that all three models fail most frequently in the 30-digit condition. 
This suggests that cognitive load increases for the models as the numerical magnitude grows.

\section{Discussion}
\subsection{Limitation of Resources Promotes LMs' Rational Inference}
Although the patterns varied across LMs, our data demonstrate that several models showed a tendency to adopt more rational comprehension strategies under the dual-task situations. This suggests that one of the reasons for errors that prioritize plausibility information over function words is the reduction of cognitive resources available for sentence comprehension. 

Could it be possible that this plausibility-based shift in the dual task is due to jumbled and longer input? We find that dependence on plausibility increased in the dual task condition compared to the noisy single task condition, even though both included identical arithmetic expressions. Thus, the effect cannot be attributed solely to input complexity. Instead, it reflects the additional cognitive demands imposed by performing two tasks simultaneously, which deplete available resources. 

Thus, an additional task is key to distinguishing the present study from \citet{Asami2024What}. They manipulate memory load by lengthening sentences: while longer sentences reduced overall accuracy, they did not alter the influence of plausibility. Therefore, our results suggest that task-induced cognitive load, rather than input length alone, is a critical factor in constraining LMs' cognitive resources.
Human working memory is not simply about storage, but about the dynamic interaction between memory access and processing \cite{Atkinson1971Control, Baddeley1974Working, Baddeley2003Working, Just1992Capacity}. Therefore, our findings suggest that LMs, like humans, rely not only on short- or long-term memory stores, but also on a working-memory-like mechanism that integrates memory with ongoing computation.

Our results also align with findings from reasoning studies. 
LMs show heuristic reasoning strategies, called ``shortcut solutions'' \cite{Geirhos2020Shortcut,Jia2017Adversarial,Ko2020Look,Tang2023Large}. They sometimes rely on superficial letter sequences rather than the content of the documents, similar to this study. 
Furthermore, it has been observed that when cognitive resources become limited, reasoning processing shifts from syntactic interpretation to more superficial comprehension based on word-level associations  \cite{Lampinen2024Language, Zhang2024Working} and degrades some functions, such as safety mechanisms 
\cite{Upadhayay2025Working,Xu2024Cognitive}. 
Our work further explores what functions are reduced in sentence comprehension. In particular, our data show that adding an arithmetic computation task depletes the resources and consequently reduces syntactic processing.

\subsection{Contributions to Psycholinguistics}
Finally, we discuss how our results can contribute to psycholinguistic theories. Our findings suggest that limiting cognitive resources induces a shift toward rational inference, i.e., a human-like comprehension strategy. More broadly, these results support the hypothesis that human-like sentence understanding fundamentally arises from how limited cognitive resources are allocated.

In human sentence comprehension, behavioral effects are often attributed to memory limitations \citep[and others]{Gibson1998Linguistic,VanDyke2003Distinguishing}. Working memory, particularly its temporary storage used for ongoing processing, has long been considered a crucial factor. However, it remains an open question whether human comprehension behavior can truly be explained solely in terms of working memory capacity.

This study provides supporting evidence from non-human systems, namely LMs, that their behavior under constrained conditions resembles human behavior. Namely, under limited cognitive resources, LMs and humans (i) reduced reliance on syntactic function, with increased reliance on their primary world knowledge \cite{Ayasse2021Principle,Ferreira2003Misinterpretation,Gibson2013Rational}, and (ii) a bias toward acquiescence \cite{Condon2006Personality,Knowles1999Why, Lechner2015Cognitive,Rammstedt2023Di}. That is, both humans and LMs degrade certain functions supporting syntactic processing and rejection when cognitive resources are constrained. As a result, they rely more on semantic plausibility.

Taken together, our results suggest that the underlying principle of human sentence comprehension may lie in the resource limitation of working memory, leading to strategies adopted to achieve efficient sentence comprehension (cf. Cognitive Load Theory by \citet{Sweller1988Cognitive} and \citet{Sweller2019Cognitive}).

\section{Conclusion}
We implement a dual-task paradigm in which the models simultaneously solve arithmetic problems and answer comprehension questions.
GPT-4o, o3-mini, and o4-mini shift their comprehension strategies under cognitive resource limitations toward rational inference, similar to humans.
Our findings suggest that task-induced cognitive load, rather than input length alone, constrains LMs' cognitive resources and makes them more human-like.

\section*{Limitations} 

First, further research is needed to explore what drives LMs’ rational reading strategies under dual-task conditions. While some models demonstrate a clear shift toward plausibility-based comprehension, others do not. Notably, even within the GPT-4 family, GPT-4o exhibits rational inference behavior, whereas GPT-4.1 does not, indicating that identifying the source of this difference is not straightforward. It remains unclear whether these variations arise from differences in internal architecture, training procedures, or other aspects of the model.\footnote{We examined whether LMs acquire a rational inference strategy during training using OLMo \cite{Groeneveld2024OLMo}, which provides access to both training data and intermediate training checkpoints. However, even the final models fail to solve the math problems in the dual-task condition.} Recent studies have proposed that attention mechanisms in LMs serve as a working-memory–like component in the context of modeling sentence processing costs \cite{Ryu2021Accounting, Timkey2023Language, Yoshida2025If}. Thus, future research could extend this line of inquiry to comprehension strategies, potentially providing new insights into human working memory processes during comprehension.

The second limitation of this study is that the results may depend on the choice of baseline and prompt design. Although our Noisy Single Task is designed to isolate the effect of adding a calculation task by comparing the Dual Task, other baselines or formatting choices may yield different outcomes. This is relevant because large LMs are known to be sensitive to prompt formulation \cite{Kojima2022Large,Schmidt2024Towards,Sclar2024Quantifying}. Thus, systematically varying prompts, including non-semantic interruptions such as delimiters or formatting tokens instead of numerics, would be an important direction for future work.

Finally, direct human–LM comparisons would be highly valuable for more detailed modeling of human sentence comprehension. Differences between humans and LMs may be sensitive to task design or stimulus properties. Importantly, establishing a robust and well-characterized dual-task paradigm for LMs is a necessary prerequisite for meaningful human–LM comparison, as premature comparisons risk conflating methodological artifacts with cognitive effects. Our contribution should therefore be viewed as a first step: introducing a dual-task paradigm for LMs and demonstrating its potential to reveal rational inference behavior.

Taken together, future work should therefore enhance the dual-task paradigm by exploring alternative baselines and prompt designs. After establishing robust experimental settings, subsequent work can pursue more detailed human–LM comparisons and deeper investigation into the internal mechanisms underlying working memory in LMs.

\section*{Ethical Considerations}
We use crowdsourcing in our human data collection. Before crowdworkers accept participation, we inform them of the purpose of the study, the tasks, the time required, the risks, the voluntary nature, the compensation, and the privacy considerations.
There are no known serious risks while participating in the experiment. Participants may experience mild fatigue from reading text on a screen for an extended period. We therefore allow them to take breaks at any time.
The experimental materials do not contain any offensive content. 
We do not obtain any personally identifiable information about participants, except for Prolific ID (used only for compensation payments), age, and sex. 
This experiment was approved by the ethics committee of the author's institution (National Institute of Informatics).

\section*{Acknowledgments}
We would like to thank the anonymous ARR reviewers for their valuable comments. 
This study was supported by JSPS KAKENHI (No. JP23KJ0199, JP25K21281), JST BOOST (No. JPMJBY24D9), and JST FOREST (No. JPMJFR232R).

\bibliography{references}

\appendix
\section{Example Stimuli for the Noisy Single Task and the Dual Task}\label{sec:stimulibyarith}
Table \ref{tab:stimulibyarith} represents example stimuli for the noisy single task and the dual task.

\begin{table*}[h]
\centering
\begin{tabular}{p{2.5cm}p{11cm}}
\toprule
\multicolumn{1}{l}{Calculation} & \multicolumn{1}{l}{Stimuli}   \\ 
\midrule
1dig.2add.                      & The 5 cocktail +
  blended 6 the = bartender x5633 and 9 the + authorities 3 agitated = the
  x5634 organist 6 after + the 8 infantryman = saluted x5635 the 3 pollster.                                \\[5pt]
1dig.3add.                      & The 7 cocktail +
  blended 6 the + bartender 5 and = the x4210 authorities 2 agitated + the 4
  organist + after 4 the = infantryman x4211 saluted 4 the + pollster.                                    \\[5pt]
3dig.2add.                      & The 952 cocktail +
  blended 604 the = bartender x5633 and 793 the + authorities 271 agitated =
  the x5634 organist 770 after + the 832 infantryman = saluted x5635 the 531
  pollster.                \\[5pt]
3dig.3add.                      & The 212 cocktail +
  blended 260 the + bartender 341 and = the x4210 authorities 220 agitated +
  the 631 organist + after 753 the = infantryman x4211 saluted 264 the +
  pollster.                    \\
\bottomrule
\end{tabular}
\caption{Example stimuli by arithmetic problem. Context stimuli in the noisy single task and the dual task embed arithmetic problems in sentences.}
\label{tab:stimulibyarith}
\end{table*}

\section{Prompts}\label{sec:prompts}
The method to make prompts is as follows. We put instructions of tasks and few-shot examples on the system message, and a sentence and a question on the user message. 
The few-shot example contains four sets of sentence and question, each of these are plausible or implausible, and question whose answer is Yes or No.
Examples of prompts are in Figures~\ref{fig:promptsingle}--\ref{fig:promptdual}.

\begin{figure}[h]
  \begin{center}
  \includegraphics[width=\columnwidth]{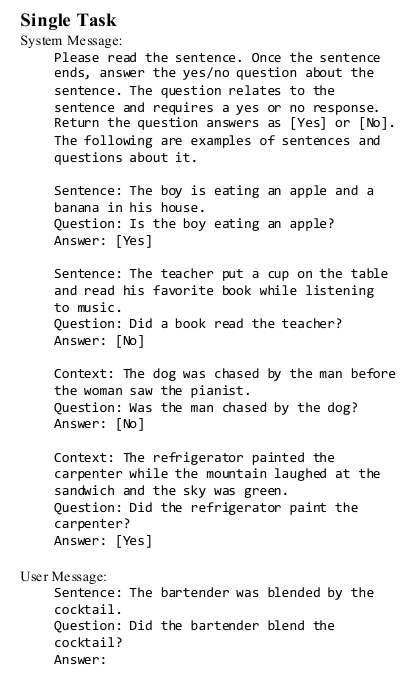}
  \caption{Example prompt of the single task.}
  \label{fig:promptsingle}
  \end{center}
\end{figure}

\begin{figure}[h]
  \begin{center}
  \includegraphics[width=\columnwidth]{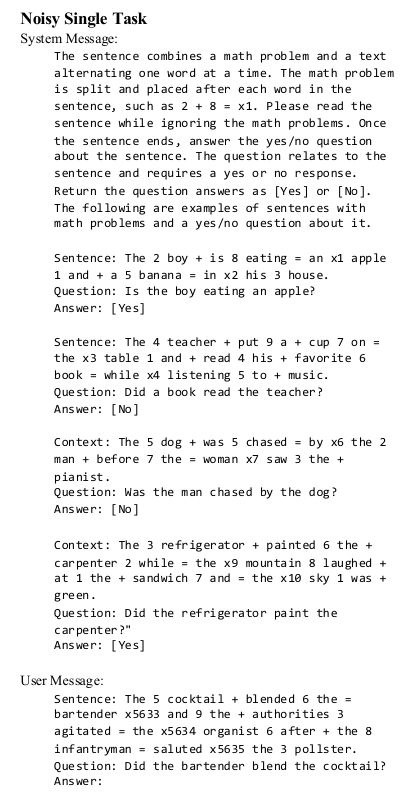}
  \caption{Example prompt of the noisy single task.}
  \label{fig:promptnoisy}
  \end{center}
\end{figure}

\begin{figure}[h]
  \begin{center}
  \includegraphics[width=\columnwidth]{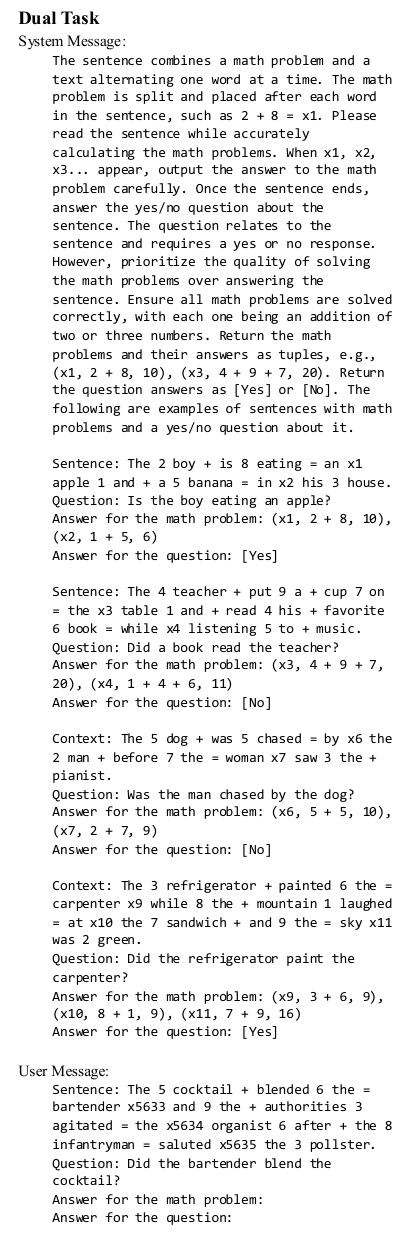}
  \caption{Example prompt of the dual task.}
  \label{fig:promptdual}
  \end{center}
\end{figure}

\section{Effect of Plausibility by Calculation}\label{sec:accbyarith}
Figure~\ref{fig:graphbyarith} represents the mean accuracy rates of comprehension questions by plausibility, LM, and calculation. The graph shows the consistent results with the results by construction and correct answer (see Figure~\ref{fig:graphbycond} and \ref{fig:graphbyanswer}): GPT-4o, o3-mini, and o4-mini have the greater difference between plausible and implausible conditions in the dual task than the single task or the noisy single task. DeepSeek-V3 also represents this trend. 

\begin{figure*}[h]
  \begin{center}
   \includegraphics[width=0.88\textwidth,keepaspectratio]{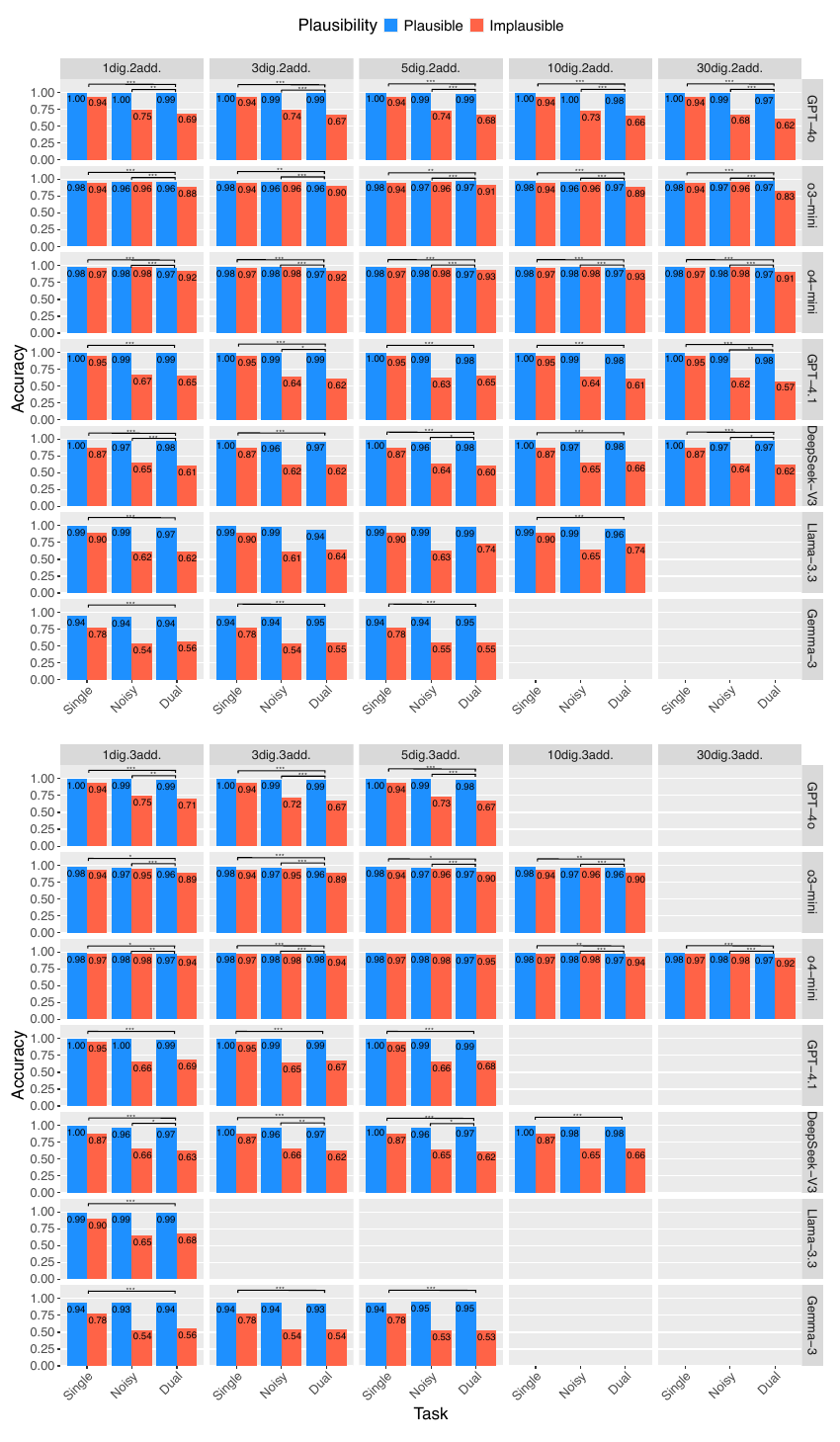}
  \caption{Accuracy rate of comprehension tasks by plausibility, task, LM, and calculation. Some conditions are excluded during preprocessing (see Section \ref{sec:preprocessing}). 
  *\textit{p} < 0.05. **\textit{p} < 0.01. ***\textit{p} < 0.001.}
  \label{fig:graphbyarith}
  \end{center}
\end{figure*}

\section{Accuracy for Calulation}
The accuracy of arithmetic problems is displayed in Table~\ref{tab:math}. The table shows that most of the LMs have around 90\% accuracy except for Llama-3.3, which has below 80\% in any conditions.

\begin{table}[h]
\centering
\scriptsize
\begin{tabular}{cccc} 
\toprule
Model       & Calculation & Mean & SD    \\ 
\midrule
GPT-4o      & 1dig.2add.  & 0.99 & 0.12  \\
GPT-4o      & 1dig.3add.  & 1.00 & 0.00  \\
GPT-4o      & 3dig.2add.  & 0.99 & 0.11  \\
GPT-4o      & 3dig.3add.  & 1.00 & 0.00  \\
GPT-4o      & 5dig.2add.  & 0.99 & 0.10  \\
GPT-4o      & 5dig.3add.  & 0.99 & 0.11  \\
GPT-4o      & 10dig.2add.  & 0.92 & 0.26  \\
GPT-4o      & 10dig.3add.  & 0.50 & 0.50  \\
GPT-4o      & 30dig.2add.  & 0.77 & 0.42  \\
GPT-4o      & 30dig.3add.  & 0.04 & 0.19  \\ \midrule
o3-mini     & 1dig.2add.  & 0.97 & 0.17  \\
o3-mini     & 1dig.3add.  & 0.98 & 0.14  \\
o3-mini     & 3dig.2add.  & 0.95 & 0.22  \\
o3-mini     & 3dig.3add.  & 0.98 & 0.14  \\
o3-mini     & 5dig.2add.  & 0.95 & 0.22  \\
o3-mini     & 5dig.3add.  & 0.97 & 0.17  \\
o3-mini     & 10dig.2add.  & 0.93 & 0.25  \\
o3-mini     & 10dig.3add.  & 0.93 & 0.25  \\
o3-mini     & 30dig.2add.  & 0.71 & 0.45  \\
o3-mini     & 30dig.3add.  & 0.58 & 0.49  \\ \midrule
o4-mini     & 1dig.2add.  & 0.97 & 0.16  \\
o4-mini     & 1dig.3add.  & 0.95 & 0.21  \\
o4-mini     & 3dig.2add.  & 0.97 & 0.17  \\
o4-mini     & 3dig.3add.  & 0.94 & 0.23  \\
o4-mini     & 5dig.2add.  & 0.96 & 0.20  \\
o4-mini     & 5dig.3add.  & 0.93 & 0.25  \\
o4-mini     & 10dig.2add.  & 0.94 & 0.24  \\
o4-mini     & 10dig.3add.  & 0.91 & 0.29  \\
o4-mini     & 30dig.2add.  & 0.82 & 0.38  \\
o4-mini     & 30dig.3add.  & 0.76 & 0.43  \\ \midrule
GPT-4.1     & 1dig.2add.  & 1.00 & 0.00  \\
GPT-4.1     & 1dig.3add.  & 1.00 & 0.05  \\
GPT-4.1     & 3dig.2add.  & 1.00 & 0.00  \\
GPT-4.1     & 3dig.3add.  & 1.00 & 0.01  \\
GPT-4.1     & 5dig.2add.  & 1.00 & 0.04  \\
GPT-4.1     & 5dig.3add.  & 0.99 & 0.11  \\ 
GPT-4.1     & 10dig.2add.  & 0.92 & 0.28  \\
GPT-4.1     & 10dig.3add.  & 0.12 & 0.33  \\
GPT-4.1     & 30dig.2add.  & 0.64 & 0.48  \\
GPT-4.1     & 30dig.3add.  & 0.03 & 0.18  \\ \midrule
DeepSeek-V3 & 1dig.2add.  & 0.99 & 0.11  \\
DeepSeek-V3 & 1dig.3add.  & 0.99 & 0.09  \\
DeepSeek-V3 & 3dig.2add.  & 0.99 & 0.12  \\
DeepSeek-V3 & 3dig.3add.  & 0.99 & 0.08  \\
DeepSeek-V3 & 5dig.2add.  & 0.97 & 0.17  \\
DeepSeek-V3 & 5dig.3add.  & 0.99 & 0.10  \\
DeepSeek-V3 & 10dig.2add.  & 0.90 & 0.30  \\
DeepSeek-V3 & 10dig.3add.  & 0.84 & 0.37  \\
DeepSeek-V3 & 30dig.2add.  & 0.91 & 0.29  \\
DeepSeek-V3 & 30dig.3add.  & 0.58 & 0.49  \\ \midrule
Llama-3.3   & 1dig.2add.  & 0.82 & 0.38  \\
Llama-3.3   & 1dig.3add.  & 0.73 & 0.44  \\
Llama-3.3   & 3dig.2add.  & 0.76 & 0.43  \\
Llama-3.3   & 3dig.3add.  & 0.50 & 0.50  \\
Llama-3.3   & 5dig.2add.  & 0.78 & 0.42  \\
Llama-3.3   & 5dig.3add.  & 0.42 & 0.49  \\
Llama-3.3   & 10dig.2add.  & 0.66 & 0.47  \\
Llama-3.3   & 10dig.3add.  & 0.26 & 0.44  \\
Llama-3.3   & 30dig.2add.  & 0.24 & 0.43  \\
Llama-3.3   & 30dig.3add.  & 0.01 & 0.08  \\ \midrule
Gemma-3     & 1dig.2add.  & 0.89 & 0.32  \\
Gemma-3     & 1dig.3add.  & 0.82 & 0.39  \\
Gemma-3     & 3dig.2add.  & 0.93 & 0.26  \\
Gemma-3     & 3dig.3add.  & 0.83 & 0.38  \\
Gemma-3     & 5dig.2add.  & 0.81 & 0.39  \\
Gemma-3     & 5dig.3add.  & 0.72 & 0.45  \\
Gemma-3     & 10dig.2add.  & 0.56 & 0.50  \\
Gemma-3     & 10dig.3add.  & 0.44 & 0.50  \\
Gemma-3     & 30dig.2add.  & 0.18 & 0.38  \\
Gemma-3     & 30dig.3add.  & 0.00 & 0.00  \\
\bottomrule
\end{tabular}
\caption{Accuracy of arithmetic problems. Some conditions are excluded during preprocessing (see Section \ref{sec:preprocessing}). SD = standard deviation. }
\label{tab:math}
\end{table}

\section{Human Experiment Procedure}\label{HumExpProc}
Procedures are almost the same as LM’s experiment. 
We conducted the experiment using PCIbex Farm (\url{https://farm.pcibex.net/}).
Instructions for each task are shown in Figures \ref{fig:instructionsingle}, \ref{fig:instructionnoisy}, and \ref{fig:instructiondual}. A task is specified first. Specifically, the instruction ``Read the sentence,'' ``Read the Sentence and IGNORE the Math Problem,'' or ``Read the Sentence and CALCULATE the Math Problem'' appeared on the screen in the single-tasks, noisy-single-tasks, or dual-tasks, respectively. Then, participants read a sentence word by word. Only in the dual task, type the answer for the arithmetic problem once x1, x2, x3… appears. When the sentence presentation ends, participants answer the comprehension question. Before the experiment, participants practice the task for four trials. 

\begin{figure*}[htbp]
    \begin{tabular}{cc}
      \begin{minipage}[t]{0.5\textwidth}
        \centering
        \fbox{\includegraphics[width=0.97\textwidth]{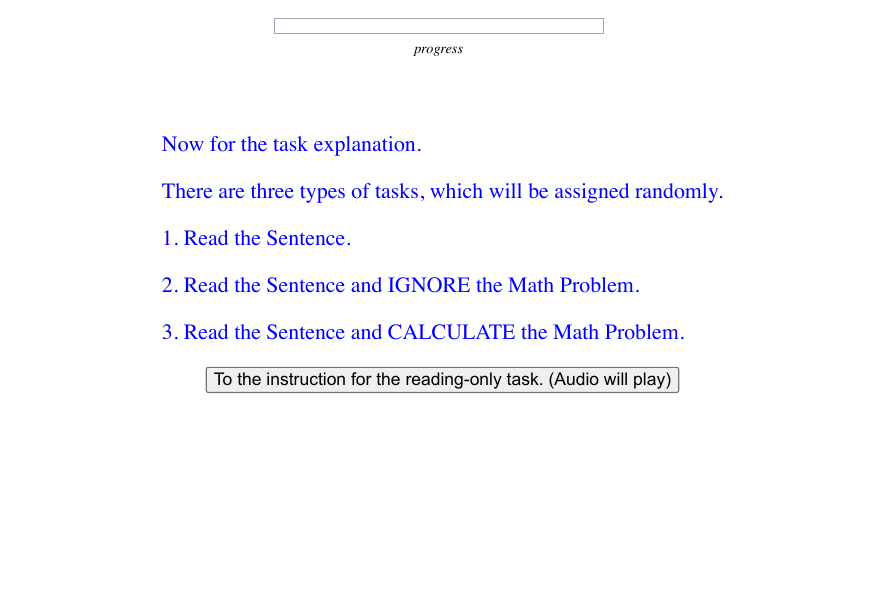}}
        \subcaption{First instruction}
      \end{minipage} 
      \begin{minipage}[t]{0.5\textwidth}
        \centering
        \fbox{\includegraphics[width=0.97\textwidth]{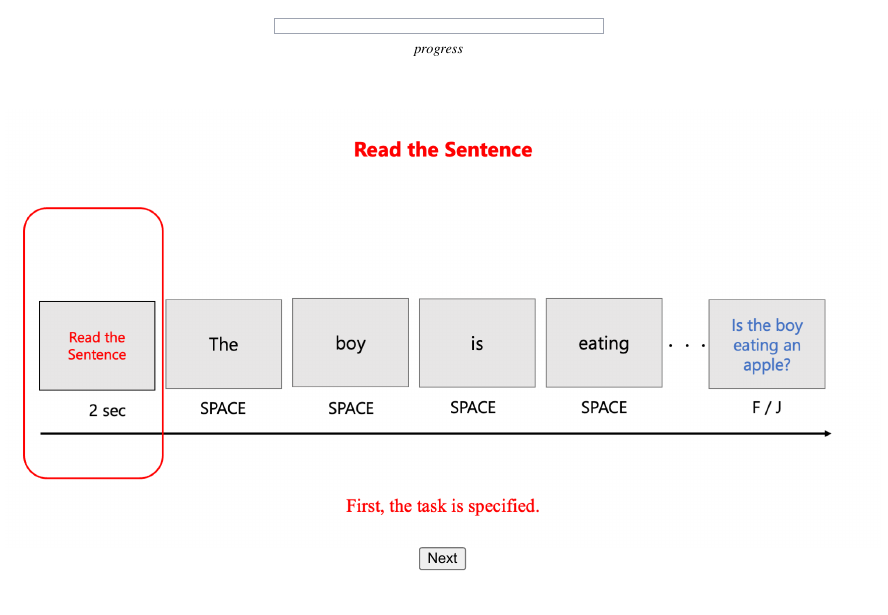}}
        \subcaption{Second instruction}
      \end{minipage} \\
      
      \vspace{3mm}
      
      \begin{minipage}[t]{0.5\textwidth}
        \centering
        \fbox{\includegraphics[width=0.97\textwidth]{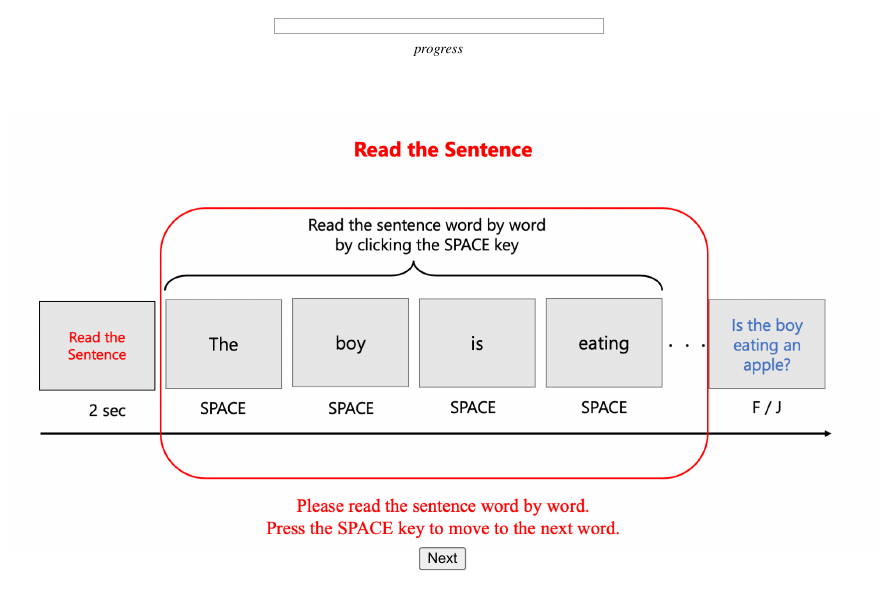}}
        \subcaption{Third instruction}
      \end{minipage} 
      \begin{minipage}[t]{0.5\textwidth}
        \centering
        \fbox{\includegraphics[width=0.97\textwidth]{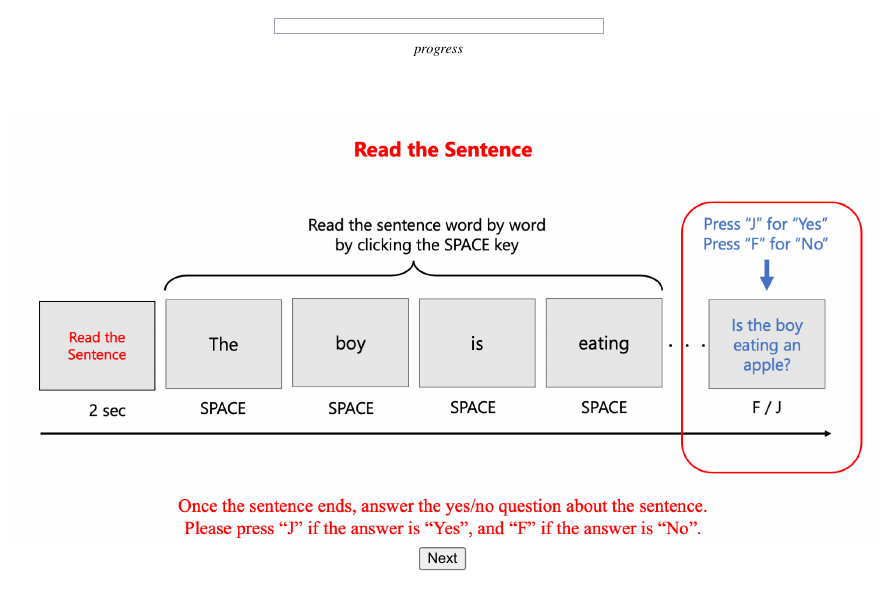}}
        \subcaption{Fourth instruction}
      \end{minipage} \\

      \vspace{3mm}

      \begin{minipage}[t]{0.5\textwidth}
        \centering
        \fbox{\includegraphics[width=0.97\textwidth]{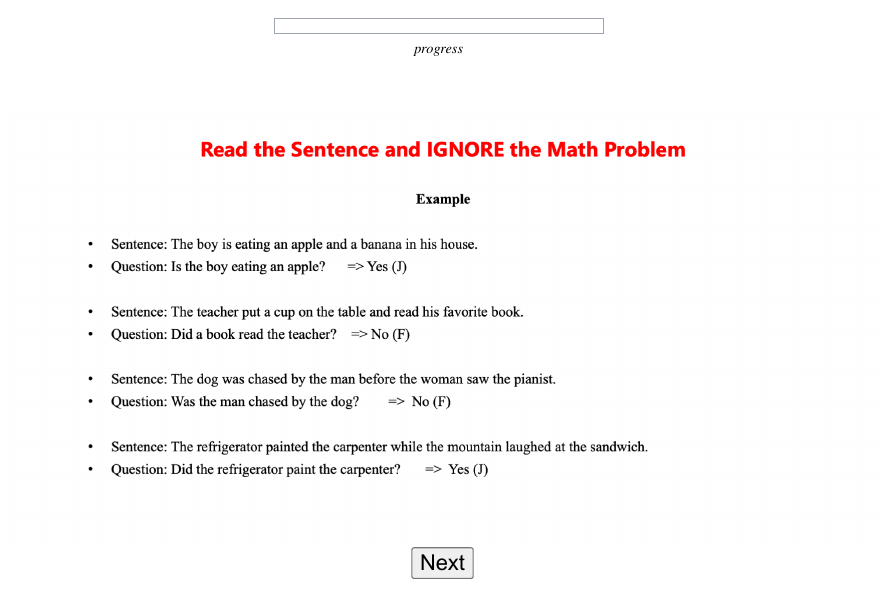}}
        \subcaption{Fifth instruction}
      \end{minipage} 
    \end{tabular}
     \caption{Instruction screens for the single task in the human experiment.}
 \label{fig:instructionsingle} 
  \end{figure*}

\begin{figure*}[htbp]
    \begin{tabular}{cc}
      \begin{minipage}[t]{0.5\textwidth}
        \centering
        \fbox{\includegraphics[width=0.97\textwidth]{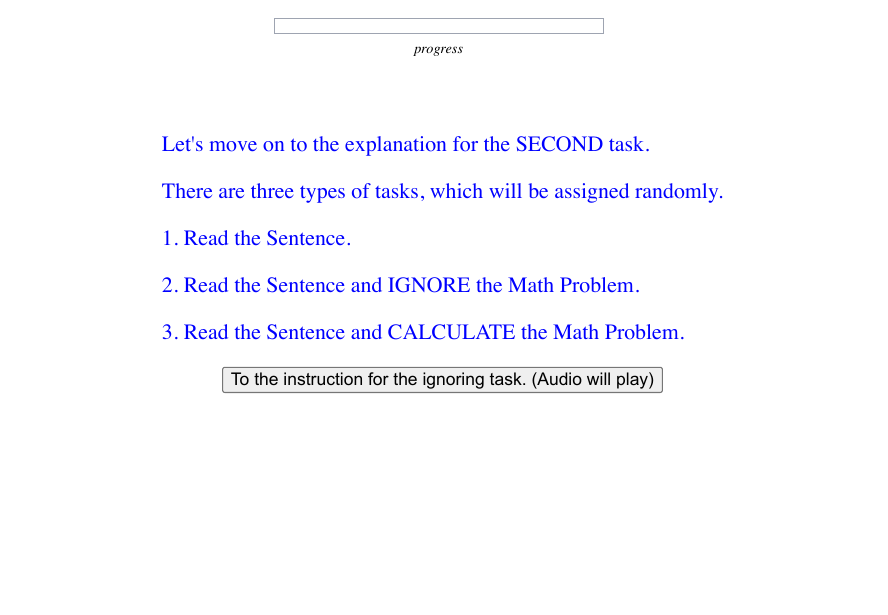}}
        \subcaption{First instruction}
      \end{minipage} 
      \begin{minipage}[t]{0.5\textwidth}
        \centering
        \fbox{\includegraphics[width=0.97\textwidth]{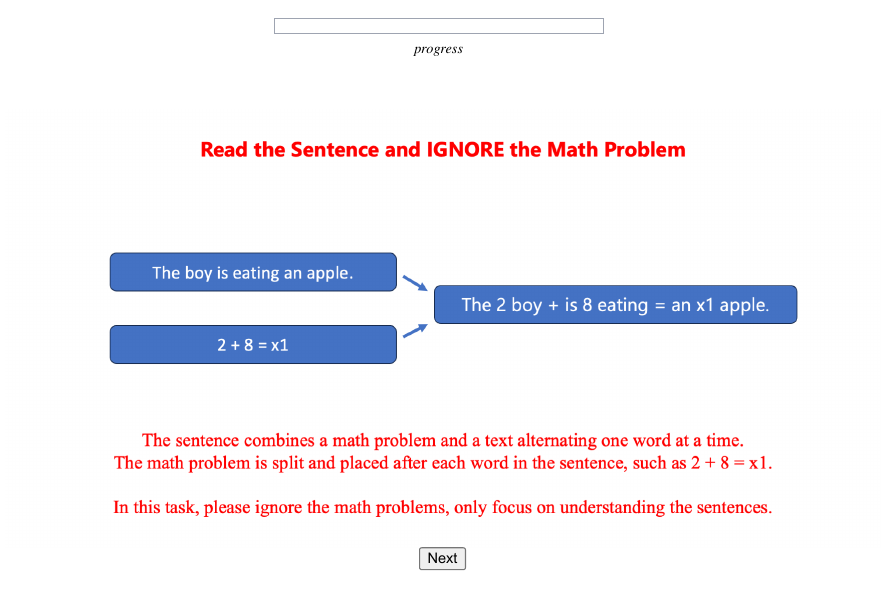}}
        \subcaption{Second instruction}
      \end{minipage} \\
      
      \vspace{3mm}
      
      \begin{minipage}[t]{0.5\textwidth}
        \centering
        \fbox{\includegraphics[width=0.97\textwidth]{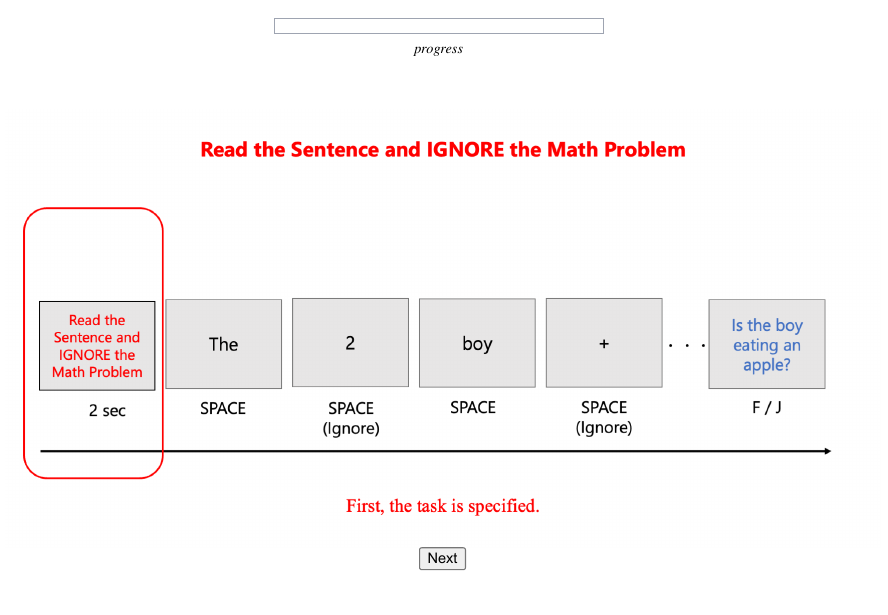}}
        \subcaption{Third instruction}
      \end{minipage} 
      \begin{minipage}[t]{0.5\textwidth}
        \centering
        \fbox{\includegraphics[width=0.97\textwidth]{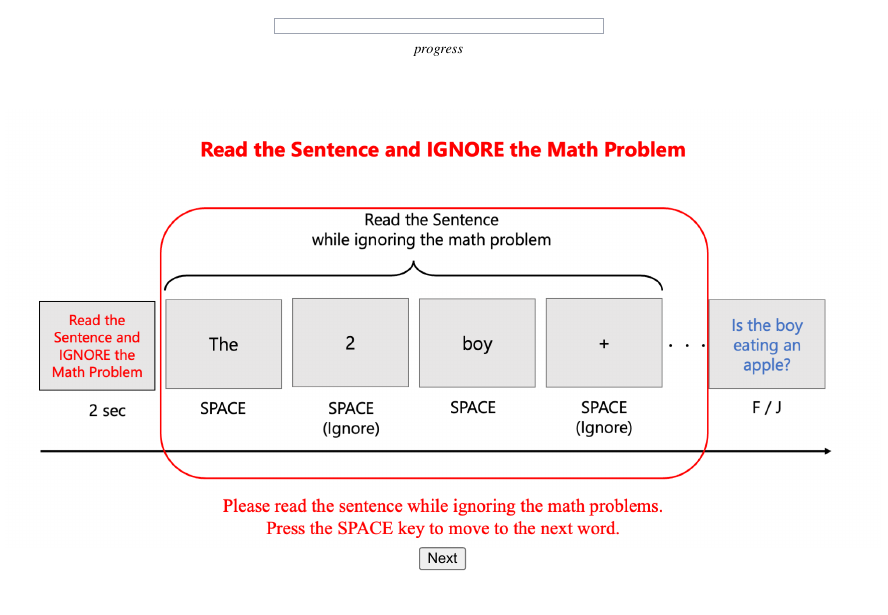}}
        \subcaption{Fourth instruction}
      \end{minipage} \\

      \vspace{3mm}

      \begin{minipage}[t]{0.5\textwidth}
        \centering
        \fbox{\includegraphics[width=0.97\textwidth]{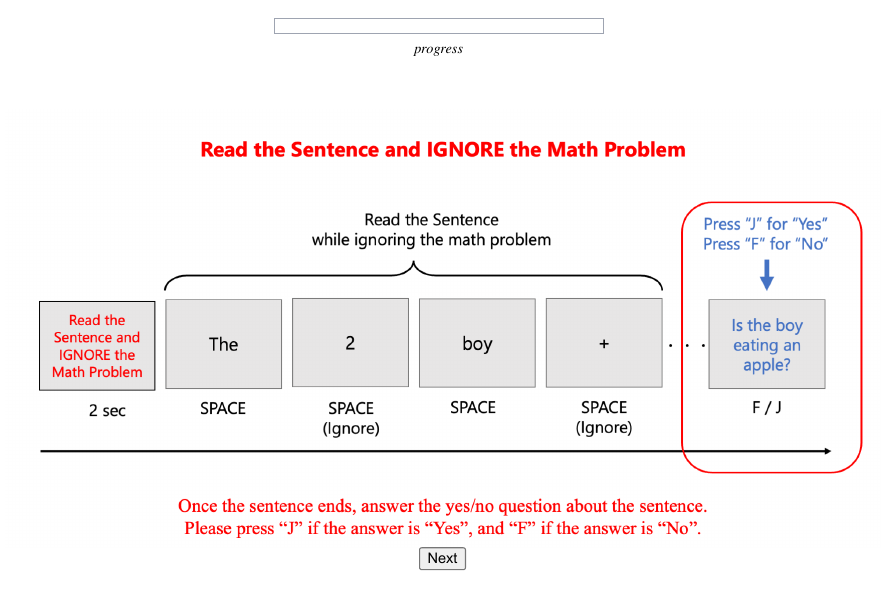}}
        \subcaption{Fifth instruction}
      \end{minipage} 
      \begin{minipage}[t]{0.5\textwidth}
        \centering
        \fbox{\includegraphics[width=0.97\textwidth]{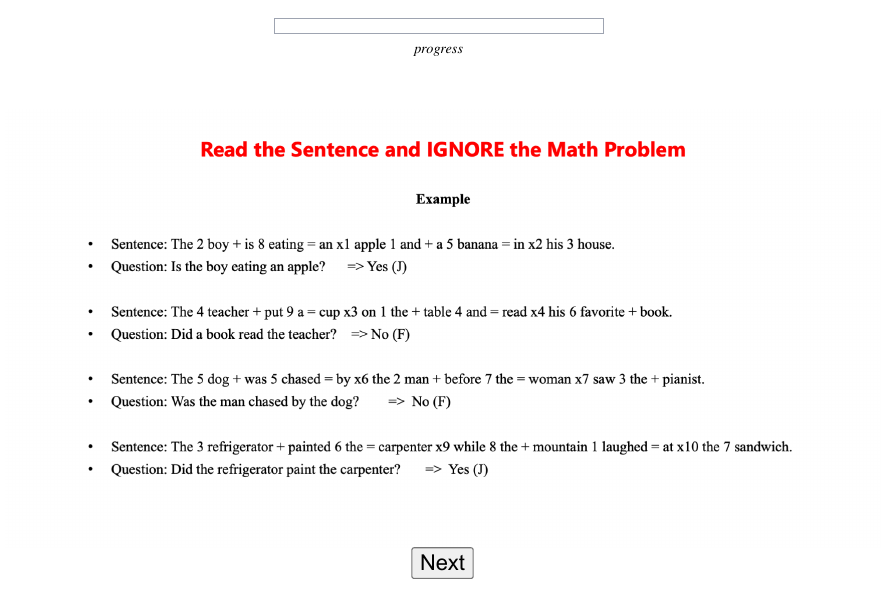}}
        \subcaption{Sixth instruction}
      \end{minipage} 
    \end{tabular}
     \caption{Instruction screens for the noisy single task in the human experiment.}
  \label{fig:instructionnoisy}
  \end{figure*}

\begin{figure*}[htbp]
    \begin{tabular}{cc}
      \begin{minipage}[t]{0.5\textwidth}
        \centering
        \fbox{\includegraphics[width=0.97\textwidth]{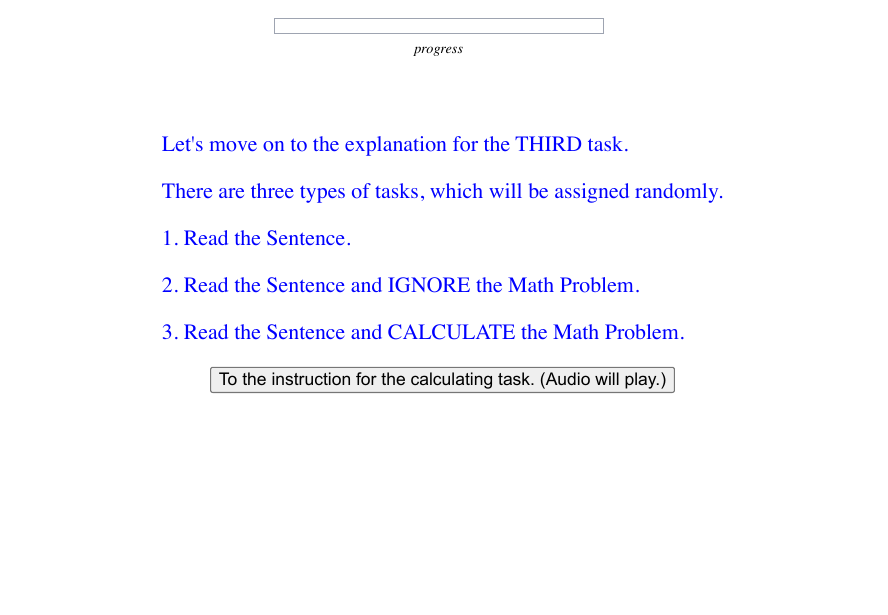}}
        \subcaption{First instruction}
      \end{minipage} 
      \begin{minipage}[t]{0.5\textwidth}
        \centering
        \fbox{\includegraphics[width=0.97\textwidth]{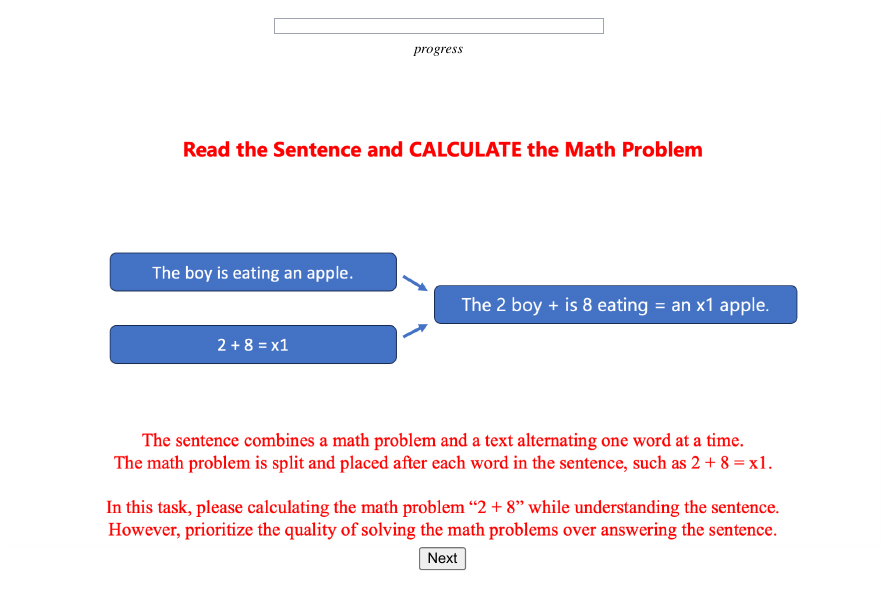}}
        \subcaption{Second instruction}
      \end{minipage} \\
      
      \vspace{3mm}
      
      \begin{minipage}[t]{0.5\textwidth}
        \centering
        \fbox{\includegraphics[width=0.97\textwidth]{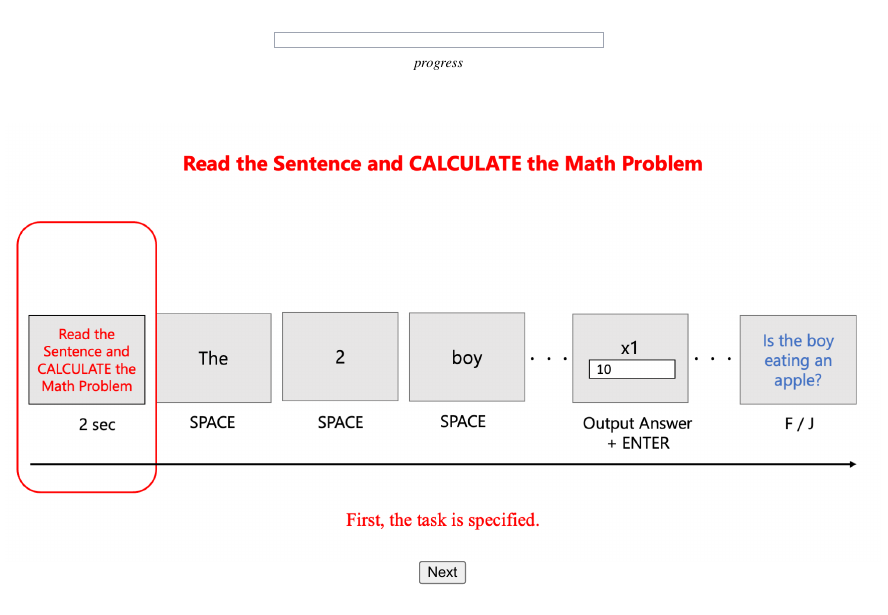}}
        \subcaption{Third instruction}
      \end{minipage} 
      \begin{minipage}[t]{0.5\textwidth}
        \centering
        \fbox{\includegraphics[width=0.97\textwidth]{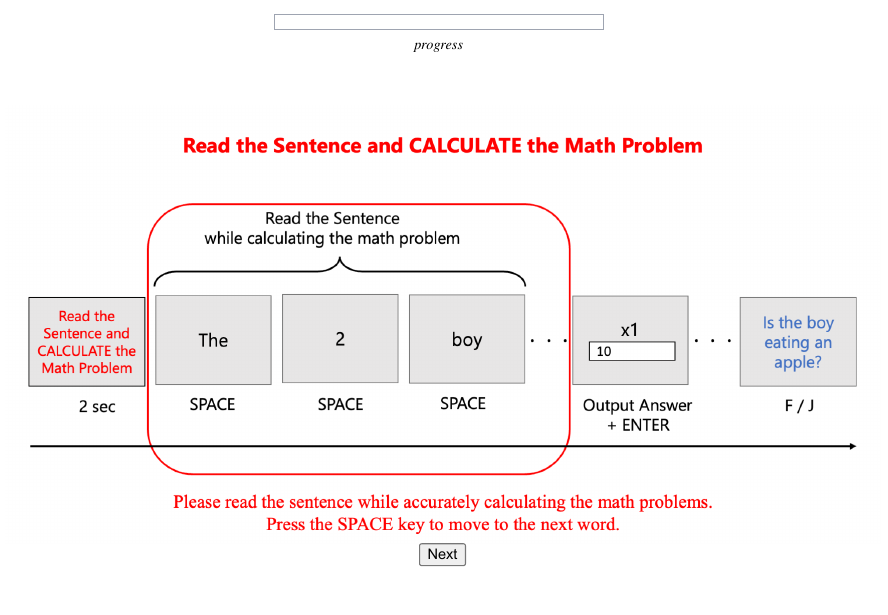}}
        \subcaption{Fourth instruction}
      \end{minipage} \\

      \vspace{3mm}

      \begin{minipage}[t]{0.5\textwidth}
        \centering
        \fbox{\includegraphics[width=0.97\textwidth]{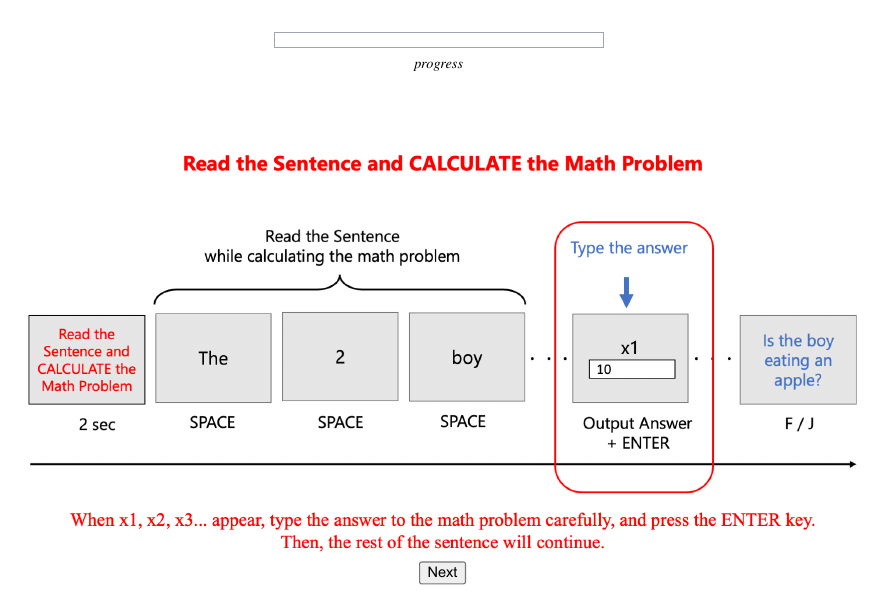}}
        \subcaption{Fifth instruction}
      \end{minipage} 
      \begin{minipage}[t]{0.5\textwidth}
        \centering
        \fbox{\includegraphics[width=0.97\textwidth]{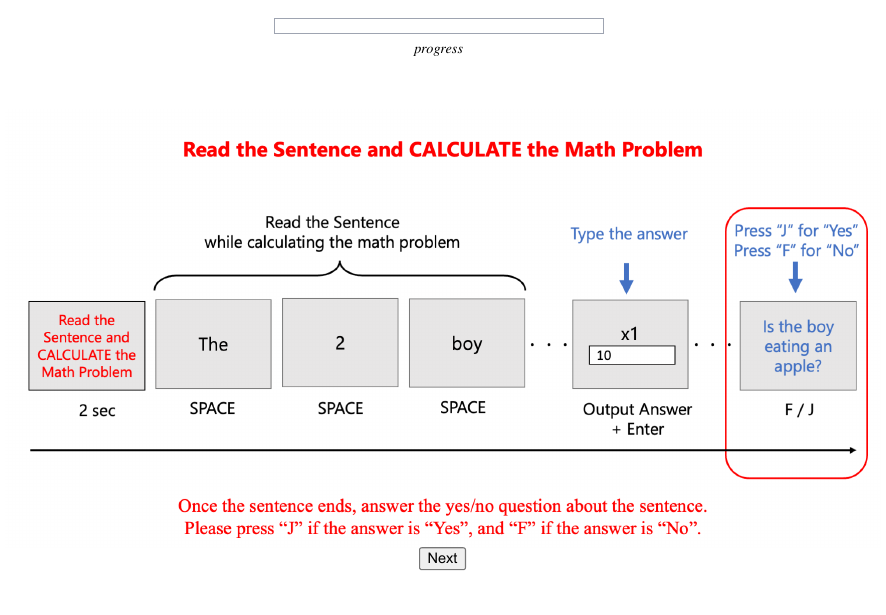}}
        \subcaption{Sixth instruction}
      \end{minipage} \\
      
      \vspace{3mm}
      
      \begin{minipage}[t]{0.5\textwidth}
        \centering
        \fbox{\includegraphics[width=0.97\textwidth]{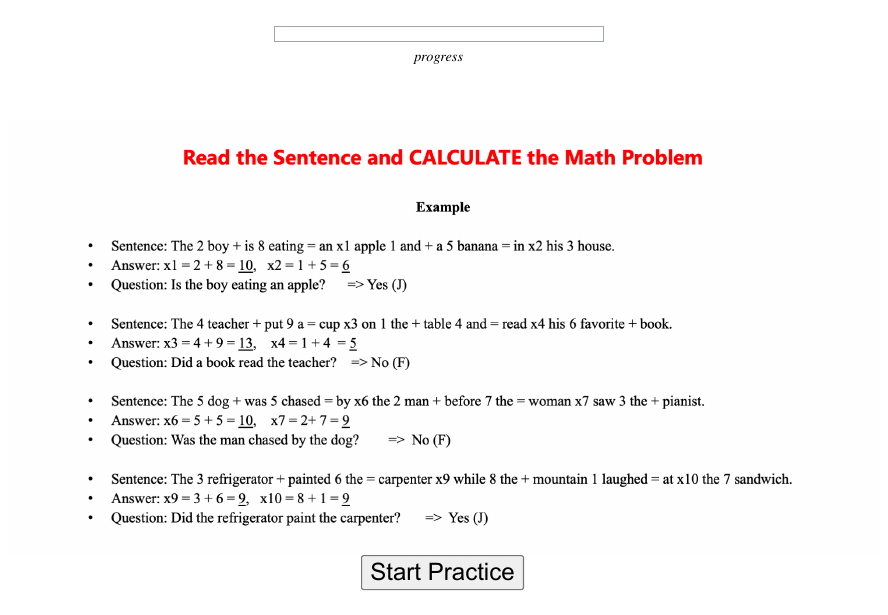}}
        \subcaption{Seventh instruction}
      \end{minipage} 
    \end{tabular}
     \caption{Instruction screens for the dual task in the human experiment.}
  \label{fig:instructiondual}
  \end{figure*}

\end{document}